\pdfoutput=1

\documentclass[11pt]{article}

\usepackage[]{ACL2023}

\usepackage{times}
\usepackage{latexsym}

\usepackage[T1]{fontenc}

\usepackage[utf8]{inputenc}

\usepackage{inconsolata}
\usepackage{enumitem}
\usepackage{hyperref}
\usepackage{color}
\usepackage{url}
\usepackage{adjustbox}
\usepackage{makecell}
\usepackage{tabularx}
\usepackage[T1]{fontenc}
\usepackage{xcolor}
\usepackage{booktabs}

\usepackage{microtype}
\usepackage{amsmath}
\usepackage{amsfonts}
\usepackage{amssymb}
\usepackage{pifont}
\usepackage{subfig}
\usepackage{graphicx}
\usepackage{booktabs}
\usepackage{multirow}
\usepackage{bm}
\usepackage{csquotes}
\usepackage{xcolor,colortbl}

\newcommand{\wikidata}{\textsc{WikiData}}
\newcommand{\OP}{\textsc{LMCRAWL}}
\newcommand{\secref}[1]{Sec.~\ref{#1}}

\title{Crawling The Internal Knowledge-Base of Language Models}

\author{Roi Cohen$^1$~~~~Mor Geva$^2$\thanks{\hspace{5px} Now at Google Research.}~~~~Jonathan Berant$^1$~~~~Amir Globerson$^1$\\ 
$^1$Blavatnik School of Computer Science, Tel Aviv University~~~$^2$Allen Institute for AI\\
\small{\texttt{roi1@mail.tau.ac.il, pipek@google.com, joberant@cs.tau.ac.il, gamir@tauex.tau.ac.il}}\\
}

\begin{document}
\maketitle

\begin{abstract}
Language models are trained on large volumes of text, and as a result their parameters might contain a significant body of factual knowledge. Any downstream task performed by these models implicitly builds on these facts, and thus it is highly desirable to have means for representing this body of knowledge in an interpretable way. However, there is currently no mechanism for such a representation.
Here, we propose to address this goal by extracting a knowledge-graph of facts from a given language model. 
We describe a procedure for ``crawling'' the internal knowledge-base of a language model. Specifically, given a seed entity, we expand a knowledge-graph around it. The crawling procedure is decomposed into sub-tasks, realized through specially designed prompts that control for both precision (i.e., that no wrong facts are generated) and recall (i.e., the number of facts generated). We evaluate our approach on graphs crawled starting from dozens of seed entities, and show it yields high precision graphs (82-92\%), while emitting a reasonable number of facts per entity.
\end{abstract}

\section{Introduction}

\begin{figure*}
\centering
\includegraphics[scale=0.52]{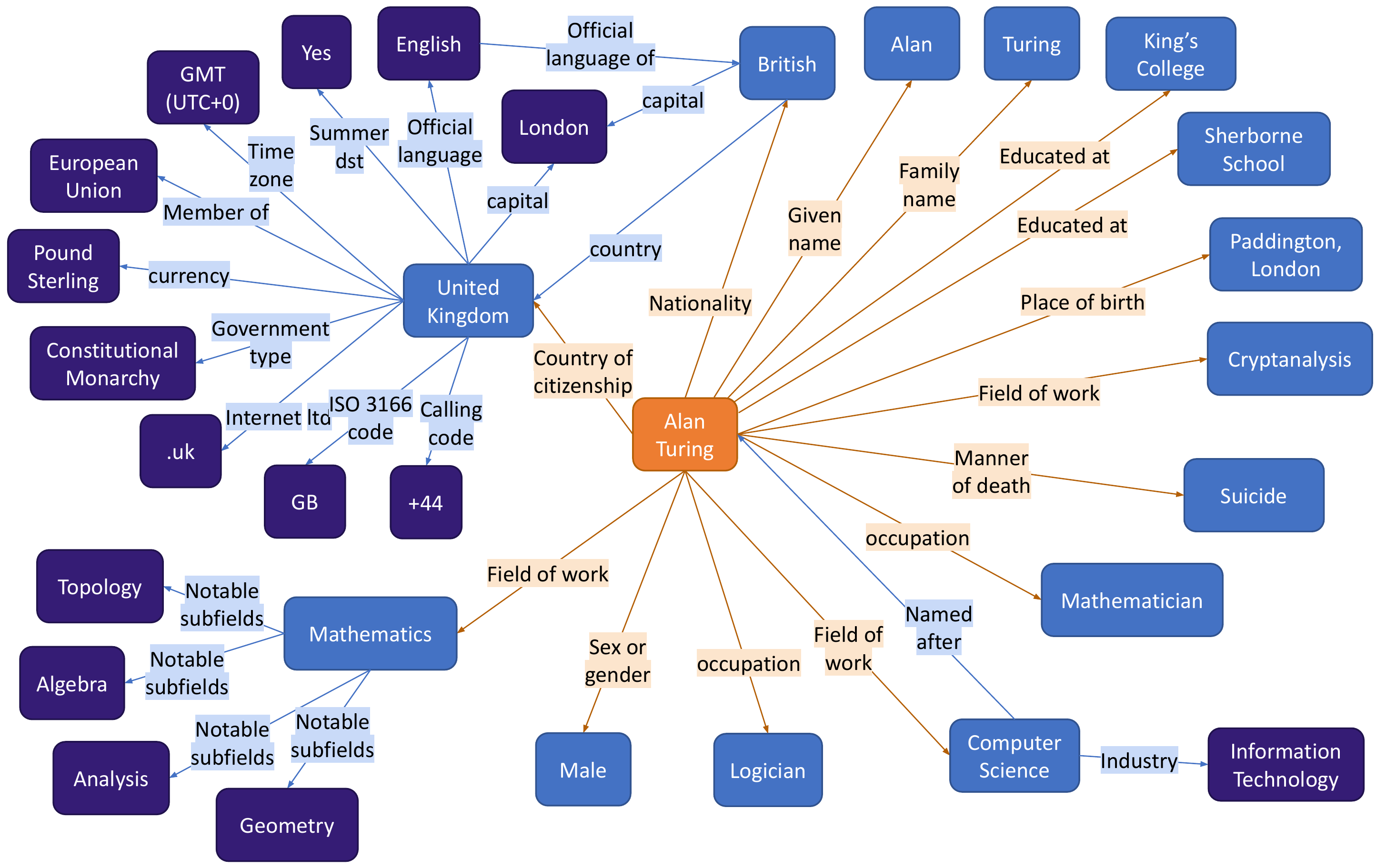}
\caption{An example of a generated depth-2 knowledge graph around the seed entity \textsc{Alan Turing}, applying \OP{} (see \secref{sec:crawling}-\secref{sec:setup}). Additional graphs are in \secref{apx:example_graphs}.}
\label{fig:knolwedge_graph}
\end{figure*}

Modern language models (LMs)  \cite{raffel2020t5,brown2020gpt3} are trained on vast amounts of text that captures much of human knowledge, including scientific articles, Wikipedia, books, and other sources of information \cite{pile}. Consequently, such models
encode world knowledge in their parameters, allowing them to generate rich and coherent outputs.

Past work has illustrated LMs can be viewed as knowledge-bases \cite{lama} as well as analyzed the encoded knowledge
\citep[e.g., see][]{alkhamissi2022review} and leveraged it for applications such as closed-book QA \cite{closed-book_qa,brown2020gpt3} and search \cite{tay2022transformer}, illustrating LMs can be viewed as knowledge-bases \cite{lama}.
But what are the facts stored in the internal knowledge bases of modern LMs, and how can these be represented explicitly? This is the challenge we address in this work. 
Our motivation is to obtain an interpretable and transparent representation that will allow humans to inspect what the LM knows, what it does not know, why it makes certain mistakes, and what are the biases it encodes. Moreover, with such a representation, one can leverage general-purpose tools, such as query languages, for interacting with this knowledge.




The first question in this endeavour is what is a suitable explicit knowledge representation.
A natural candidate structure is a knowledge graph (KG). Namely, a graph whose nodes are entities and whose edges represent relations between entities.
KGs are appealing since information can be readily ``read-off'' from the graph, they can be reliably queried, and different KGs can be easily compared.
KGs have been extensively used to represent knowledge \cite{bollacker2008freebase,vrandevcic2014wikidata},
but a key limitation is their \emph{low coverage}, as they usually require manual curation
and depend on a closed schema.
Conversely, LMs might have very high coverage as they are trained on a vast body of knowledge represented as raw text.
We thus ask if it is possible to convert an LM to a KG, such that we enjoy its advantages while achieving high coverage.

As the full KG encoded in an LM can be large, we reduce the problem to the task of constructing a KG around a given seed entity. For example, Fig.~\ref{fig:knolwedge_graph} shows a KG extracted by our method for the seed entity \emph{Alan Turing}. This can be viewed as a crawling procedure which starts from the seed entity and recursively expands it to expose additional facts. 
This crawling problem introduces several new challenges. First, unlike prior work \cite{lama,prompting_as_probing, bertnet}, 
we are given only \emph{an entity}, without knowing what relations are associated with it.
Thus, we have to extract those relations and then find the 
objects for each relation.
Second, KGs are expected to exhibit very high precision, and thus it is necessary to generate as many relevant facts as possible while maintaining almost perfect factual correctness.\footnote{We note that there is a deeper philosophical aspect to this issue, which is at the core of the field of epistemology. Namely, what does it mean for a model to ``believe'' a fact, as opposed to the model ``knowing'' a fact. Here we adopt a ``dispositional'' view of belief, whereby a belief corresponds to a statement by the model, and knowledge is a belief that is true in the world.}

We address the above challenges by decomposing crawling into multiple sub-tasks, and handle each task using few-shot in-context learning \cite{brown2020gpt3}.
Explicitly, we do not fine-tune a model, but instead manually design a prompt and a few examples for each task, an approach recently-proven successful \cite{wei2022chain,drozdov2022compositional,chowdhery2022palm,khot2022decomposed}.
We use the following sub-tasks (see Tab. \ref{tab:prompts} for the full list and examples). 
First, given an entity $e$ (e.g., \textsc{Alan Turing}), we generate the relations relevant for $e$ (e.g., \textsc{Educated at}, \textsc{Place of Birth}). Second, for each entity $e$ and relation $r$, we generate the corresponding set of objects $O$ and add to the KG triplets $(e, r, o)$ for each $o \in O$. For example, for \textsc{Alan Turing} and \textsc{Educated at}, we generate triplets with the \emph{objects} \textsc{King's college} and \textsc{Sherborne School}. To maintain high precision, we prompt the model to emit \emph{``Don't know''} whenever it is not confident about the target objects. All the above outputs are generated through in-context learning, where we use the \wikidata{} KG \cite{WikiData} to construct in-context examples. \emph{Don't know} examples are constructed by finding
true facts in \wikidata{} that are unknown to the LM. Finally, we increase recall by prompting the LM to generate paraphrases for entities and relations, and use those to obtain additional triplets.


We test our approach with GPT-3 (\texttt{text-davinci-002}) on 140 seed entities, and show that we can extract accurate KGs ($\sim$82-92\% precision) that contain a plausible
number of facts per entity.
Importantly, large LMs are not constrained to a predefined schema, and indeed our procedure with GPT-3 generates facts outside the schema of \wikidata{}, e.g., \textsc{(Boston Celtics, Championships, 17)}.

To conclude, our contributions are: 1) Formulating the problem of crawling a KG from an LM, 2) Presenting 
a prompt-based approach that decomposes the problem into multiple sub-tasks, and 3) Evaluating the approach with GPT-3, which leads to 
high-precision graphs.



\section{Problem Setup}
Our goal is to uncover the knowledge-base of a given LM. We represent a knowledge-base via a KG, which is a collection of triplets. Formally, a KG is a graph $G = (N, R, E)$, where $N$ is a set of entities, $R$ is a set of relations, and $E$ is a set of subject-relation-object triplets $(s,r,o)$ where $s,o\in N$ and $r\in R$. 

To simplify the setup, we assume we are given a ``seed entity'' around which we will expand the graph (for example Fig.~\ref{fig:knolwedge_graph}). Conceptually, we can also let the LM generate seed entities, but we argue seed expansion is a more realistic scenario, where a user is interested in a graph about a certain entity.



Entities and relations are represented via strings and are not constrained to a given vocabulary \citep[similar to open information extraction. e.g., see][]{vo2017open}.

\section{Crawling KGs via Prompting}
\label{sec:crawling}

\begin{table*}[ht]
\setlength\tabcolsep{4.0pt}
\footnotesize
\begin{center}
\begin{tabular}{ p{0.105\linewidth} p{0.125\linewidth}  p{0.55\linewidth}  p{0.15\linewidth} }
\textbf{Sub-task} & \textbf{Query} & \textbf{Prompt} & \textbf{Expected Output} \\ 
\toprule
Relation \mbox{Generation}  & \texttt{\textcolor{teal}{Philippines}}
& \texttt{\textcolor{olive}{\textbf{Q}: René Magritte \textbf{A}: ethnic group, place of birth, place of death, sex or gender, spouse, country of citizenship, member of political party,  native language, place of burial,  cause of death, residence, family name, given name, manner of death, educated at, field of work, work location, represented by \textbf{Q}: Stryn \textbf{A}: significant event, head of government, country, capital, separated from}
\textcolor{teal}{\textbf{Q}:~Philippines \textbf{A}:}}
& \texttt{\textcolor{purple}{leader name \# cctld \# capital \# calling code}}  \\
\midrule
Pure Object Generation  &  \texttt{\textcolor{teal}{Barack Obama \# child}}
& \texttt{\textcolor{olive}{\textbf{Q}: Monte Cremasco \# country \textbf{A}: Italy \textbf{Q}: Johnny Depp \# children \textbf{A}: Jack Depp \# Lily-Rose Depp \textbf{Q}: Wolfgang Sauseng \# employer \textbf{A}: University of Music and Performing Arts Vienna} 
    \textcolor{teal}{
    \textbf{Q}: Barack Obama \# child \textbf{A}:
    }}
& \texttt{\textcolor{purple}{Sasha Obama \# Malia Obama}}  \\
\midrule
DK Object Generation & \texttt{\textcolor{teal}{Queen Elizabeth II \# date of death}}
& \texttt{\textcolor{olive}{\textbf{Q}: Heinrich Peters \# occupation \textbf{A}: Don't know \textbf{Q}: Monte Cremasco \# country \textbf{A}: Italy \textbf{Q}: Ferydoon Zandi \# place of birth \textbf{A}: Don't know \textbf{Q}: Hans Ertl \# sport \textbf{A}: mountaineering}
    \textcolor{teal}{\textbf{Q}: Queen Elizabeth II \# date of death \textbf{A}:}} 
& \texttt{\textcolor{purple}{Don't know}} \\
\midrule
Subject \mbox{Paraphrasing} &  \texttt{\textcolor{teal}{Alan Turing}}
& \texttt{\textcolor{teal}{Alan Turing is also known as:}}
& \texttt{\textcolor{purple}{The father of computing}} \\
\midrule
Relation Paraphrasing  & \texttt{\textcolor{teal}{notable work}}
& \texttt{\textcolor{teal}{'notable work' may be described as}}
&  \texttt{\textcolor{purple}{a work of 'great value' or a work of 'importance'}} \\
\bottomrule
\end{tabular}
\end{center}
\caption{The full list of sub-tasks in our approach, where for each sub-task we provide its name, a query, a corresponding prompt, and the expected output. 
In `DK Object Generation' the prompt declares in one of the in-context examples that the model does not know the place of birth of Ferydoon Zandi, since querying for it leads to a wrong answer (the query with the wrong answer isn't shown).}

\label{tab:prompts}
\end{table*}

\begin{figure*}
\begin{center}
\includegraphics[scale=0.5]{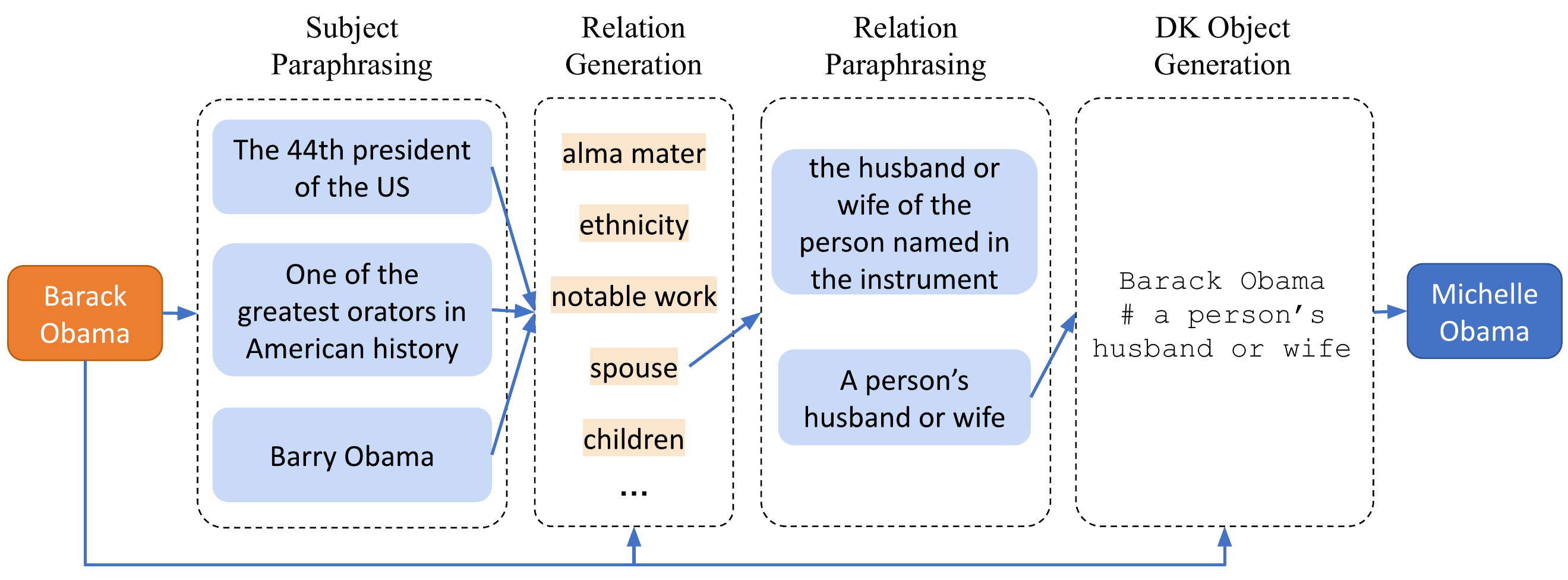}
\end{center}
\caption{An illustration of the full method for crawling a subgraph (\OP{}), starting from \textsc{Barack Obama} as the subject, until obtaining the triplet \textsc{(Barack Obama, spouse, Michelle Obama)}.}
\label{fig:full_method_illustration}
\end{figure*}

The core component of our approach is a procedure that takes an entity $e$, and extracts all relations associated with it, and the corresponding objects. Namely, we expand the KG around this entity. We can then recursively apply this procedure to further expand the KG. We refer to this as `entity expansion', and break it into two high-level steps:
\begin{itemize}
[leftmargin=*,topsep=2pt,parsep=2pt]
    \item \textbf{Relation generation} (Sec.~\ref{sec:relation_gen}): For an entity $e$, generate a set of relations $R$, where $e$ is the subject.
    \item \textbf{Object generation} (Sec.~\ref{sec:object_gen}-Sec.~\ref{sec:dontknow}): Given the entity $e$ and the relation set $R$, find the corresponding objects. Namely, for each $r \in R$, find a list of entities $O$ such that $(e,r,o)$ is in the KG for $o \in O$. We consider lists since many relations  (e.g., $\mbox{\textsc{Children}}$) potentially have multiple correct objects. Furthermore, we also consider the case where the object corresponding to $(e,r)$ is unknown to the model (e.g., the model does not know who is the daughter of a given entity $e$). In this case we take $O$ to be empty, and the edge is not added to the KG. This is crucial for maintaining a high-precision KG.
\end{itemize}
Both steps are achieved via few-shot in-context learning. Namely, we construct prompts with in-context examples (stay fixed throughout the process) that exhibit the desired behaviour (Tab.~\ref{tab:prompts}).


To improve recall, we employ an additional paraphrasing procedure (Sec.~\ref{sec:relation_augment} and Sec.~\ref{sec:object_augment}), 
which generates alternative strings for a given entity or relation. For example, the entity \textsc{William Clinton} can be referred to as \textsc{William Jefferson Clinton} or \textsc{Bill Clinton}, and the relation \textsc{Occupation} may be expressed as \textsc{Profession}.
Thus, we run object and relation generation for all these variants,
and pool the results to construct the final graph. Paraphrases are also obtained through the LM, without use of external knowledge. The entire flow is illustrated in Fig.~\ref{fig:full_method_illustration}, and we next elaborate on each of the components.

\subsection{Relation Generation \label{sec:relation_gen}}
Our task is to generate a set of relations $R$ for a given subject entity $e$. To achieve this, we leverage \wikidata{} to construct in-context examples.
Specifically, we pick a list of \wikidata{} entities $e_1,\ldots, e_{K_r}$ and for each entity $e_i$, extract its set of \wikidata{} relations. This results in $K_r$
in-context examples for relation generation. We concatenate the target entity to the in-context examples, feed this prompt to the LM and use its output as the set $R$ for $e$.
Tab.~\ref{tab:prompts} shows an example prompt.
We note that this generation process can produce relations that are not included in the prompt, and are not part of \wikidata{} at all.\footnote{For example, when the subject is a sports team, the model repeatedly generated a relation regarding its \textsc{mascot} or \textsc{largest win}, which are facts outside of \wikidata{}.} Full prompt with in-context examples is presented in Sec.~\ref{appen:relation_prompt}.

\subsection{Relation Paraphrasing \label{sec:relation_augment}}
A relation $r$ may be described in multiple ways, and the LM might work better with some of these paraphrases \cite{jiang-etal-2021-know}. Thus, we use a procedure to obtain a set of paraphrases of $r$, denoted by $P(r)$, and run all downstream crawling tasks for all strings in $P(r)$.

For relation paraphrasing we find that in-context examples are not necessary and an instruction prompt is sufficient.
Tab.~\ref{tab:prompts} shows a specific example under the sub-task ``Relation Paraphrasing''.
See Sec.~\ref{relation_para_extended_details} for the three prompts and more technical details. 

\subsection{Object Generation \label{sec:object_gen}}
Our next goal is, for each $r\in R$, to generate a set of objects $O$ such that $(e,r,o)$ is in the KG for all $o\in O$. Importantly, we should also let the LM declare it does not know the object, and thus $O$ would be empty. In this case, no edge will be added to the output KG.


We first explain prompt construction without the use of \emph{``Don't Know''} output, and refer to this as ``Pure Object Genration''. 
We take $K_o$ entities $e_1,\ldots,e_{K_o}$ from \wikidata{}. For each entity $e_i$, we choose one of its relations $r_i$, and all the objects $O_i$ for this entity-relation pair in \wikidata{}. This creates $K_o$ examples for object generation.
Similar to relation generation, the target entity-relation pair is concatenated to the $K_o$ examples, and the list of objects is parsed from the generated LM output (see exact format in Tab.~\ref{tab:prompts}, under the sub-task ``Pure Object Generation'', and the full prompt with in-context examples in Sec.~\ref{appen:pure_obj_prompt}).
Recall that for each relation, we have multiple paraphrases. To maintain high precision, we only accept objects that were generated by at least two realizations of the relation.

\subsection{Learning to Output \emph{``Don't Know''}}
\label{sec:dontknow}
A key desideratum for KGs is high precision, namely the facts in the graph should be correct with high probability.
Towards this end, we want to prompt the LM to output \emph{``Don't Know''} (DK) for facts where it is likely to make an error.\footnote{A model might make an error because it is not confident about the answer, or because its training data contains false facts. In this work, we are agnostic to this distinction and our prompt's goal is to encourage generation of correct outputs.}


But how do we know what the model does not know? To capture this, we find cases where the LM outputs erroneous facts, and use these to construct in-context examples with a DK target. For example, suppose we run `Pure Object Generation' with $e=\textsc{Bill Clinton}$ and $r=\textsc{Children}$ and the model outputs $O=\textsc{Klay Thompson}$. We deduce that the model does not know who Clinton's children are, and therefore, can add the example $e_i=\textsc{Bill Clinton}, r_i=\textsc{Children}, o_i=\emph{Don't know}$ to the prompt. 
In other words, we find examples where $o_i$ is \emph{Don't know} through cases where the model errs on its predicted objects. 
We then construct a prompt with a total of $K_{\emph{dk}}$ examples, half of which are failure cases where with $o_i=\emph{Don't know}$ and the other half are correct predictions.
We refer to this as ``DK Object Generation''. See the corresponding row in Tab.~\ref{tab:prompts} and the full prompt with in-context examples in Sec.~\ref{appen:dk_obj_prompt}.



\subsection{Subject Paraphrasing}
\label{sec:object_augment}
Similar to relations, an entity $e$ may have several names, and it may be easier for the LM to complete the triplet $(e,r,?)$ with one of these. Thus, we take a paraphrasing approach to extend an entity name $e$ into a set $P(e)$. The procedure is identical to relation paraphrasing (Sec.~\ref{sec:relation_augment}), except 
we use a single prompt 
instructing the LM to complete the sentence ``$s$ is also known as'', where $s$ is the subject.
To increase the number of paraphrases, we sample from the model three times, resulting in up to three paraphrases.

Both here and in Relation Paraphrasing (Sec.~\ref{sec:relation_augment}), the LM  occasionally generates nonsensical paraphrases. Nevertheless, the DK method handles those cases well, outputting \emph{"Don't know"} for most of them.
Thus, we argue that paraphrasing combined with DK emission is an effective approach for controlling recall and precision.




\subsection{\OP{}}
Fig.~\ref{fig:full_method_illustration} shows the application of the complete pipeline (which we refer to as \OP{}) for the entity \textsc{Barack Obama}. First, we obtain all paraphrases for $e$ (Sec.~\ref{sec:object_augment}). Then, we extract all relations for these (Sec.~\ref{sec:relation_gen}). Next, we paraphrase relations (Sec.~\ref{sec:relation_augment}). Finally, we extract the known objects for these relations (Sec.~\ref{sec:object_gen}-Sec.~\ref{sec:dontknow}).

\section{Experimental Setup}
\label{sec:setup}


As mentioned in \secref{sec:crawling}, we use \wikidata{} (publicly available) in constructing the in-context prompts. 
The number of in-context examples is $K_{r}=7$, $K_{o}=8$, $K_{\emph{dk}}=10$.

Additionally, we use \wikidata{} to select seed entities for evaluating our approach. For these seeds,  we consider the task of constructing KGs around the corresponding entities.

We split the seed entities into a validation set (20 entities), which is used to make design choices (e.g., choosing prompt format), and a test set (120 entities), which is used only for the final evaluation.

For the development set, we manually chose 20 entities from \wikidata{}. These included 
women and men with various professions, cities, countries, and various cultural entities such as movies and books. We also aimed to reprsent both head and tail entities in this list.

To construct our test set, we defined 25 specific world-entities related categories, which we refer to as the \emph{test categories}. 
Some of these were more specific, such as \emph{AI Researchers}, and some are more general, such as \emph{Scientists} (see Table.\ref{tab:main_test_set_seeds} for the full list). We chose 4 seeds out of each category as follows. We first sorted the set of entities of each group based on the number of \wikidata{} facts associated with them (we view this count as an approximate measure of popularity). Then, we randomly sampled two entities out of the full list, and an additional two out of the first 1000. Intuitively, the first two represent tail entities, while the other two represents head ones. Thus we ended up with 100 seed entities (i.e., 4 different entities out of each of the 25 different subgroups). We refer to these as the \emph{main test set} (see Tab.~\ref{tab:main_test_set_seeds}). We created an additional test set of 20 entities that is meant to contain very popular entities. 
Its entities were randomly sampled out of a set of size 1000, which was manually constructed by choosing 40 very well-known entities (i.e., that all people would know) from each of the 25 test categories.

All 140 entities were not used in the construction of any of the prompts in \secref{sec:crawling}.
Tab.~\ref{tab:val_and_head_test} shows the full list of validation and head test entities.

\begin{table}[t]
\setlength{\belowcaptionskip}{-10pt}
\footnotesize
\begin{center}
\begin{tabular}{l | l}
\textbf{Dev Seeds}  &\textbf{Head Test Seeds}
\\ \toprule
ABBA                                 &  Aristotle  \\
Alan Turing                          &  Canada  \\
Angela Merkel                        &  Celine Dion    \\
Augustin-Louis Cauchy                &  China \\
Barack Obama                         &  Emanuel Macron  \\
Bob Dylan                            &  Franz Kafka   \\
Boston Celtics                       &  Grease \\
David Bowie                          &  Hamlet   \\
Diana, Princess of Wales             &  Jacinda Ardern   \\
Eike von Repgow                      &  Lionel Messi    \\
Inglourious Basterds                 &  Little Women  \\
Marble Arch                          &  Manchester United F.C. \\
Marie Curie                          &  Margaret Hamilton  \\
Mikhail Bulgakov                     &  Michelangelo  \\ 
Moby-Dick                            &  Mike Tyson  \\
Pablo Picasso                       &   Oprah Winfrey  \\
Paris                               &   Rosalind Franklin\\
Philippines                         &   Steven Spielberg  \\
Rachel Carson                        &  Serena Williams  \\
Shahar Pe'er                         &  The Rolling Stones  \\
\bottomrule
\end{tabular}
\caption{List of all validation and head test seeds. }
\label{tab:val_and_head_test}
\end{center}
\end{table}

\paragraph{Evaluation metrics}
Given an entity $s$, our entity expansion process returns a knowledge graph $G$, that contains the entity $s$, other entities and relations between them. 

Ideally, we want to compare $G$ to a ground truth graph that results from expanding the entity $s$. Given such a graph, we could measure precision and recall over the gold and predicted sets of triplets. However, using large LMs to generate graphs leads to several challenges. 
First, there is no ground-truth graph. While we could presumably use the \wikidata{} graph, we found that it is missing many correct facts predicted by the LM. In fact, improving coverage is a key motivation for our work!
Second, facts may be reworded in several equivalent ways,
rendering comparison between \wikidata{} graphs and predicted graphs difficult.

To circumvent these challenges, we use the following notions of precision and recall.

\begin{itemize}
[leftmargin=*,topsep=2pt,parsep=2pt]
\item 
\textbf{Precision:}  
\label{subsec:precision_eval}
To estimate precision we conducted both manual and automatic evaluations (the automatic approach was more scalable). For the manual evaluation we simply tried to validate each of the generated facts by manually browsing highly trustful web sources (Google, Wikipedia, etc.) to check if the fact is true. The automatic evaluation approach was implemented as follows.
In order to check the correctness of a given predicted triplet $(e,r,o)$, we issue a query containing $(e,r)$ to Google search, and search whether $o$ appears in the result. We limit the result to first 40 words which are not HTML labels or URL links. If it does, we assume the triplet is correct. \footnote{This paragraph typically contains either an ``answer box'' or some summary of the first result page, in case there is no answer box.}
See \secref{sec:precision_underestimation} for an accuracy estimation of the automatic method.


Manual evaluation was done for all the \emph{head test set} graphs, as well as all the 1-hop graphs of the \emph{main test set}. Additionally, we performed manual evaluation for 20\% randomly sampled triplets from the 2-hop graphs (altogether, the total portion of manually labeled facts from each graph was $\sim$30\%).
The rest of the triplets were automatically evaluated.

\item 
\textbf{Recall:} 
Estimating recall is not possible since we do not have access to the true ground truth graph. Moreover, using \wikidata{} graph size as an estimate for the number of true facts will be misleading since it has low coverage in general, and \emph{high variance} in terms of coverage for different entities. Thus, we simply report the number of verified triplets in our KG. In other words, we report recall without the denominator. We refer to this as {\bf $\mbox{\# of facts}$}. This practice is similar to open information extraction \cite{vo2017open}, where it is impossible to know the set of all true facts and thus the convention is to report the number of generated facts only.


\end{itemize}

\paragraph{Implementation details}
\label{subsec:implementation_details}
As the LM in our experiments, we used the OpenAI  \texttt{text-davinci-002} model. We experiment with both greedy decoding and sampling 3 outputs per query (temperature 0.8).
We generate graphs with either a single expansion step or two expansion steps, recursively expanding entities found in the first step. After a graph is generated, 
we remove duplicates by iterating through the facts and removing a fact if the token-wise F$_1$ between it and another fact is higher than 0.85.


\paragraph{Base Model and Ablations}
The simplest version of our model includes only 'Relation Generation' (\secref{sec:relation_gen}) and 'Pure Object Generation' (\secref{sec:object_gen}), without the \emph{``Don't Know''} and paraphrasing components. We refer to this version as \emph{Pure-Greedy} and \emph{Pure-Sampling}, depending on the decoding used (see \secref{subsec:implementation_details}). 
In other model variants, we use \emph{DK} to refer to using `DK Object Generation' instead of `Pure Object Generation'. Additionally, \emph{SP} and \emph{RP} refer to adding `Subject Paraphrasing' and `Relation Paraphrasing' respectively. 

\section{Results}

\begin{table*}[t]
\setlength{\belowcaptionskip}{-10pt}
\setlength\tabcolsep{4.0pt}
\footnotesize
\begin{center}
\begin{tabular}{l | ll | ll| ll | ll|}
 \multicolumn{1}{c}{}  &\multicolumn{4}{c}{\textbf{Main Test Set}} & \multicolumn{4}{c}{\textbf{Head Test Set}} \\
 \multicolumn{1}{c}{}  & \multicolumn{2}{c}{one-hop} & \multicolumn{2}{c}{two-hop} & \multicolumn{2}{c}{one-hop} & \multicolumn{2}{c}{two-hop} \\ 
\cmidrule{2-9}
\multicolumn{1}{c|}{} & \textbf{Precision}  & \textbf{\# of Facts}  & \textbf{Precision}  & \textbf{\# of Facts} & \textbf{Precision}  & \textbf{\# of Facts}  & \textbf{Precision}  & \textbf{\# of Facts} \\
\toprule
Pure-Greedy             & $54.6 \pm 8.2$       & $ 6.2 \pm 2.8$    
                        & $43.4 \pm 6.1$       & $ 26.1 \pm 5.5$
                        & $80.3 \pm 8.4$       & $14.4 \pm 3.9$    
                        &  $62.1\pm 7.3 $       & $82.3 \pm 15.4$   \\
\OP                     & $83.3 \pm 7.9$       & $ 5.4 \pm 1.1$   
                        & $82.0 \pm  7.5$       & $ 21.4\pm 4.7$
                        & $91.5 \pm 11.4$       & $11.0 \pm 4.6 $   
                     & $90.9 \pm 4.9 $       & $61.2 \pm 25.1$      \\
\bottomrule
\end{tabular}
\end{center}
\caption{Averaged results across all 100 \textbf{main test} seeds (left), as well as all the 20 \textbf{head test} ones (right). 
}
\label{table:test_results} 
\end{table*}

We next report results showing that our expansion method is able to generate meaningful knowledge subgraphs, when expanding seed entities. 

\paragraph{Example graph:} We begin with an illustrative example for the graph of the seed entity \textsc{Alan Turing}. Fig.~\ref{fig:knolwedge_graph} shows a subset of the two-hop extracted graph in this case. It can be seen that all facts are sensible, except for the fact that the field of Computer Science is named after Alan Turing (although he is certainly one of its fathers). See also Figs.~\ref{fig:angela_markel} and ~\ref{fig:boston_celtics} for additional example graphs.

\paragraph{Results on the Main Test set:} Tab.~\ref{table:test_results} reports averaged results of the Pure-Greedy base model and \OP{} across the 100 main test seeds.
We observe that precision of Pure-Greedy is too low to be useful for a KG -- 54.6\% for 1-hop graphs and 43.4\% for 2-hop graphs. Conversely, precision with \OP{} is much higher: 83.3\% for 1-hop graphs and 82.0\% for 2-hop graphs. While we suffer a small hit in `\# of facts', the sizes of KGs output by our approach are quite reasonable.

\paragraph{Results on the Head Test set:} Tab.~\ref{table:test_results} reports averaged results of the Pure-Greedy base model and \OP{} across the 20 head test seeds.
Specifically, we achieve precision of \textbf{91.5\%} while applying \OP{} for 1-hop graphs, and for 2-hop we have \textbf{90.9\%}.
It can be seen that both precision and number of facts in this case are higher than in the main test set. This suggests that either it is easier to extract facts from the LM about popular entities, or that the LM indeed encodes more facts for these (see \secref{subsec:cov_to_freq} for further analysis).


\subsection{Ablations}
Next, we examine the contribution of each component in our final approach on the validation set.


\begin{table}[t]
\setlength\tabcolsep{4.0pt}
\footnotesize
\begin{center}
\begin{tabular}{lll}
\textbf{Method}  & \textbf{Precision}  & \textbf{\# of Facts} \\
\toprule
Pure-Sampling                        & $64.9 \pm 20.2$       & $22.2 \pm 9.7$  \\
Pure-Greedy             &$77.5 \pm 17.4$    &$12.5 \pm 6.0$  \\
\midrule 
DK-Sampling                         & $71.4 \pm 19.9$       & $17.7 \pm 9.4$   \\
DK-Greedy                        & $82.9 \pm 16.0$       & $10.2 \pm 5.9$   \\
~~ +RP      & $80.9 \pm 17.0$       & $12.7 \pm 5.4$   \\
~~ +SP      & $80.6 \pm 17.0$       & $12.2 \pm 7.0$   \\
\OP{}         & $88.3 \pm 8.2$       & $13.0 \pm 5.9$   \\
\bottomrule
\end{tabular}
\end{center}
\caption{Averaged results over the 20 \textbf{validation} seed (\textbf{one-hop}). 
DK: ``Don't know''. SP: Subject Paraphrasing. RP: Relation
Paraphrasing. 
}
\label{depth-1-full-results}
\end{table}


\begin{table}[t]
\setlength\tabcolsep{4.0pt}
\footnotesize
\begin{center}
\begin{tabular}{lll}
\textbf{Method}  & \textbf{Precision}  & \textbf{\# of Facts} \\
\toprule
Pure-Sampling                        & $40.0 \pm 9.5$     & $224.0 \pm 81.1$     \\
Pure-Greedy             & $55.9 \pm 9.7$       & $87.8 \pm 39.7$   \\
\midrule
DK-Sampling                         & $54.7 \pm 8.6$     & $144.0 \pm 83.5$    \\
DK-Greedy                        & $72.4 \pm 7.5$     & $45.8 \pm 30.3$      \\
\OP{}         & $86.4 \pm 6.1$       & $69.8 \pm 52.9$  \\
\bottomrule
\end{tabular}
\end{center}
\caption{Averaged results across all 20 \textbf{validation}  seeds (\textbf{two-hop}). DK: ``Don't know''. SP: Subject Paraphrasing. RP: Relation Paraphrasing.
}
\label{depth-2-full-results}
\end{table}

\paragraph{The Effect of Don't Know Generation:}
The goal of allowing the model to output ``Don't Know'' is to improve precision. Tab.~\ref{depth-1-full-results} and~\ref{depth-2-full-results} show results for the model without using DK prompting (in \emph{Pure} rows) as well as with (\emph{DK} rows) for both sampling and greedy decoding. In both cases, the DK option leads to much higher precision, but reduces the number of generated facts. 
However, we later recover some of these lost facts using subject and relation paraphrasing.

\paragraph{The Effect of Paraphrasing:}

Tab.~\ref{depth-1-full-results} shows results without the paraphrasing component in the \emph{DK-Greedy} row. Both paraphrasing techniques, RP and SP, separately increase coverage, while causing a minimal hit to precision.
Interestingly, combining RP and SP
leads to improvements 
in \emph{both} precision and coverage for 1-hop \emph{and} 2-hop graphs  (Tab.~\ref{depth-1-full-results},~\ref{depth-2-full-results}).


\subsection{Coverage vs. Entity Frequency}
\label{subsec:cov_to_freq}
The frequency of entities on the Web is highly skewed. That is, some entities appear many times, while others are rare. 
We expect this will be reflected in the number of facts extracted for these entities. Indeed, on \wikidata{}, head entities usually have many more facts compared to tail entities. Here, we ask whether a similar phenomenon exists in our predicted KGs. 

\begin{figure}[htp]
\begin{center}
\includegraphics[scale=0.6]{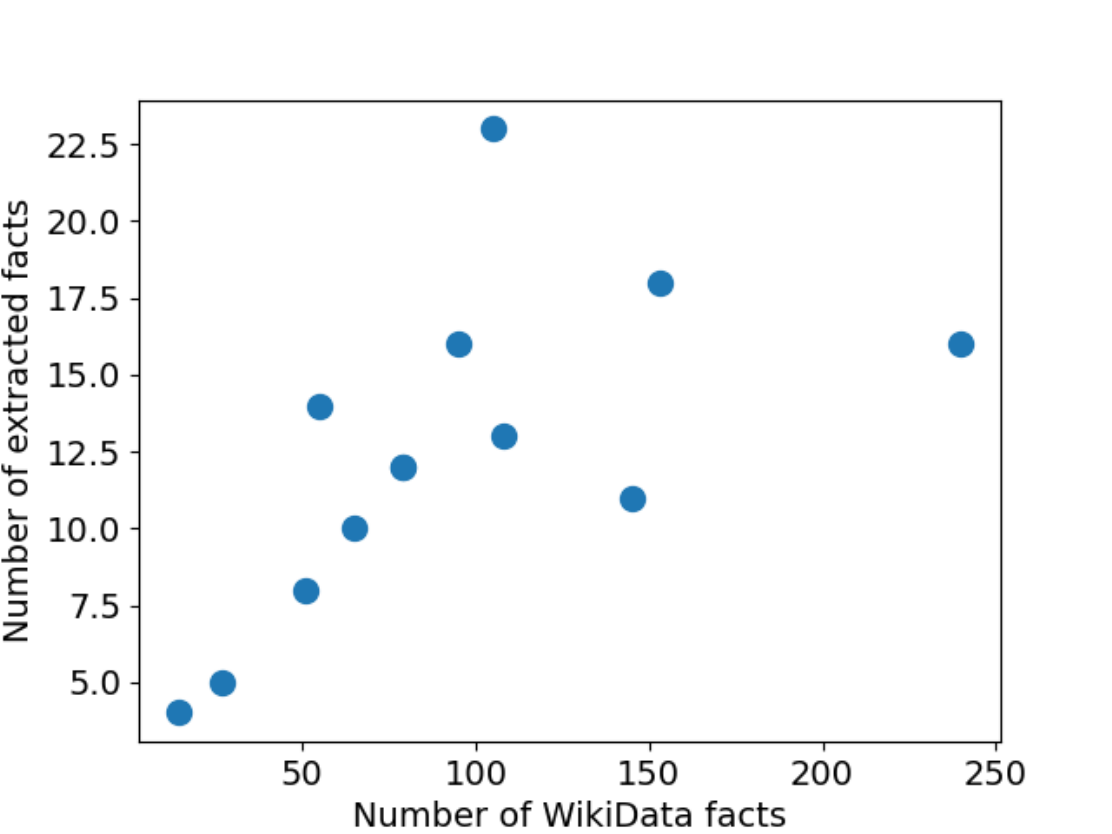}
\end{center}
\caption{The \emph{\# of triplets} extracted by \OP{} as a function of the \emph{\# of triplets} in \wikidata, for the set of validation entities of type \textsc{Person}.}
\label{absolute_recall_against_popularity}
\end{figure}

Fig.~\ref{absolute_recall_against_popularity} shows
the number of facts generated for a depth-1 graph by \OP{} for all entities of type \textsc{Person}, as a function of the number of facts that appear in the corresponding depth-1 \wikidata{} graph of the same seed. Clearly, there is high correlation (correlation coefficient is 0.61) between the number of extracted facts and entity frequency on \wikidata{}. This is rather surprising and encouraging since our procedure does not make any use of entity frequency, and head and tail entities are expanded in exactly the same way.

\subsection{Precision is Possibly Underestimated}
\label{sec:precision_underestimation}
Our automatic approach for evaluating precision uses Google search (see \secref{subsec:precision_eval}). We view this as a conservative estimate of precision, since a fact judged as true via this mechanism is highly likely to be true.
Conversely, a true fact might not be verified due to search or string matching issues. To quantify this, we sampled 500 generated facts from \emph{Pure-Greedy} and \OP{} that were judged to be incorrect through Google search, as well as 500 that were judged to be correct. We manually inspected them and found that 4.1\% of the triplets that the automatic approach has labeled as correct, are actually wrong, while 22\% of the triplets that the automatic approach has labeled to be incorrect, are true (few demonstrations are presented in \secref{apx:automatic_eval}).
Exact estimation of precision would require \emph{full} manual annotation, which we avoided to minimize costs.

\section{Related Work}


Pretrained LMs are at the heart of recent NLP research and applications. As mentioned earlier, \citet{lama} and other works have observed that LMs contain rich factual knowledge. We elaborate on other relevant works below.

\textbf{Knowledge-base construction.} 
KG construction typically involves both manual and automated aspects.
For example, popular KBs such as WordNet \cite{wordnet}, ConceptNet \cite{conceptnet} and \wikidata{} \cite{WikiData} were constructed by heavily relying on manual effort, gathering knowledge from humans.
To reduce such manual labor, automated information extraction (IE) methods have been extensively developed \cite{yates2007textrunner,fader2011identifying,paper:IE,vo2017open}. Knowledge in LMs is a fairly recent topic of interest, and has mostly focused on probing for specific facts \citep{lama, lms_as_kbs}.

Most similar to our work are
\newcite{bertnet}, who also extract KGs from LMs, 
However, they require defining the relations of interest through examples before crawling, 
while our specific goal is to start with a seed entity and allow the LM to determine the relevant relations.
Another relevant recent work is \citet{prompting_as_probing} who also use in-context learning to extract a KG from GPT3. But they also assume relations are provided, whereas a key aspect of our approach is generating the relations.

 


To the best of our knowledge, ours is the first work to construct a knowledge graph via extracting knowledge directly from LMs, using only one seed entity (and no other given relations or entities).

\textbf{Quantifying Uncertainty in LMs.} 
Factual correctness in LMs has attracted recent interest, because it is a crucial requirement for LM applicability. 
In this context, some works have studied selective question answering, where LMs avoid answering particular questions \cite{investigating_selective_predictions}. Other works have  considered calibration in LMs \cite{jiang-etal-2021-know, calib_of_pretrained}, 

Finally, recent works have investigated whether models can express their certainty on output facts, either in words or by producing the probability of certainty \cite{express_uncertainty_in_words, LM_know_what_they_know}. A key aspect of our approach is the use of a ``Don't know'' mechanism, which is related to this line of work since it lets the LM declare its certainty as part of the output. Unlike \newcite{LM_know_what_they_know}, we do so in the context of crawling a KG and via in-context learning (as opposed to fine-tuning).




\section{Conclusion}
Understanding large LMs is a key part of modern NLP, as they are used across the board in NLP applications. In particular, it is important to understand the body of knowledge these models possess, so it can be used and revised as needed, thereby avoiding factual errors and biases. In this work we present an important step towards this goal by extracting a structured KB from an LM.

There are many possible exciting extensions for our work. The first is to expand it to a larger graph corresponding to more expansion hops. This would require many more calls to an API, which at present is also costly, and it would be important to develop more cost-effective approaches. Second, we have introduced several approaches to controlling the precision and recall of the proposed model, but certainly more can be envisioned. For example, we can introduce various consistency constraints to increase precision (e.g., check that \textsc{Father of} and \textsc{Child of} are consistent in the generated graph). Finally, once a larger KG has been extracted, one can query it to see how well it serves as a question answering mechanism.

Overall, we find the possibility of seamlessly converting LMs to KGs for better interaction and control to be an exciting and fruitful direction for future research.

\section*{Limitations}

Producing the full internal KG out of an LM is still a significant challenge. One challenge is cost (as noted above). The other is error propagation issues. Once the model makes a generation mistake in a particular node of the generated graph, it may lead to an increasing number of mistakes during the next generation steps, expanding from that node. That is one of our main rationales for creating and evaluating only two-hop graphs, and not additional hops (although ideally, the real goal is to uncover the full internal KG).

Our automatic way of evaluating precision is only approximate, 
which means our reported accuracy numbers for 2-hop are an approximation of true precision (although we believe the true precision is in fact higher, as discussed in the text).

Another challenge we do not address is understanding the source of knowledge inaccuracies. Are they due to limitations of our model in extracting the knowledge, or due to the LM not containing these facts at all. This is certainly important to understand in order to improve knowledge representation in LMs.
We are also aware to the fact that since the generated graphs are not perfectly accurate, they might contain disinformation and misleading facts. That would hopefully be improved by future research. 

Finally, the question whether we could have come up with a better-reflecting ``recall'' metric than the one we suggested is yet to be solved, as in general it is still unclear how to measure knowledge coverage.

\bibliography{anthology,custom}

\begin{thebibliography}{28}
\expandafter\ifx\csname natexlab\endcsname\relax\def\natexlab#1{#1}\fi

\bibitem[{Alivanistos et~al.(2022)Alivanistos, Santamar{\'\i}a, Cochez, Kalo,
  van Krieken, and Thanapalasingam}]{prompting_as_probing}
Dimitrios Alivanistos, Selene~B{\'a}ez Santamar{\'\i}a, Michael Cochez,
  Jan-Christoph Kalo, Emile van Krieken, and Thiviyan Thanapalasingam. 2022.
\newblock Prompting as probing: Using language models for knowledge base
  construction.
\newblock \emph{arXiv preprint arXiv:2208.11057}.

\bibitem[{AlKhamissi et~al.(2022)AlKhamissi, Li, Celikyilmaz, Diab, and
  Ghazvininejad}]{alkhamissi2022review}
Badr AlKhamissi, Millicent Li, Asli Celikyilmaz, Mona Diab, and Marjan
  Ghazvininejad. 2022.
\newblock A review on language models as knowledge bases.
\newblock \emph{arXiv preprint arXiv:2204.06031}.

\bibitem[{Angeli et~al.(2015)Angeli, Premkumar, and Manning}]{paper:IE}
Gabor Angeli, Melvin Jose~Johnson Premkumar, and Christopher~D Manning. 2015.
\newblock Leveraging linguistic structure for open domain information
  extraction.
\newblock pages 344--354.

\bibitem[{Bollacker et~al.(2008)Bollacker, Evans, Paritosh, Sturge, and
  Taylor}]{bollacker2008freebase}
Kurt Bollacker, Colin Evans, Praveen Paritosh, Tim Sturge, and Jamie Taylor.
  2008.
\newblock Freebase: a collaboratively created graph database for structuring
  human knowledge.
\newblock In \emph{Proceedings of the 2008 ACM SIGMOD international conference
  on Management of data}, pages 1247--1250.

\bibitem[{Brown et~al.(2020)Brown, Mann, Ryder, Subbiah, Kaplan, Dhariwal,
  Neelakantan, Shyam, Sastry, Askell, Agarwal, Herbert-Voss, Krueger, Henighan,
  Child, Ramesh, Ziegler, Wu, Winter, Hesse, Chen, Sigler, Litwin, Gray, Chess,
  Clark, Berner, McCandlish, Radford, Sutskever, and Amodei}]{brown2020gpt3}
Tom Brown, Benjamin Mann, Nick Ryder, Melanie Subbiah, Jared~D Kaplan, Prafulla
  Dhariwal, Arvind Neelakantan, Pranav Shyam, Girish Sastry, Amanda Askell,
  Sandhini Agarwal, Ariel Herbert-Voss, Gretchen Krueger, Tom Henighan, Rewon
  Child, Aditya Ramesh, Daniel Ziegler, Jeffrey Wu, Clemens Winter, Chris
  Hesse, Mark Chen, Eric Sigler, Mateusz Litwin, Scott Gray, Benjamin Chess,
  Jack Clark, Christopher Berner, Sam McCandlish, Alec Radford, Ilya Sutskever,
  and Dario Amodei. 2020.
\newblock \href
  {https://proceedings.neurips.cc/paper/2020/file/1457c0d6bfcb4967418bfb8ac142f64a-Paper.pdf}
  {Language models are few-shot learners}.
\newblock In \emph{Advances in Neural Information Processing Systems},
  volume~33, pages 1877--1901. Curran Associates, Inc.

\bibitem[{Chowdhery et~al.(2022)Chowdhery, Narang, Devlin, Bosma, Mishra,
  Roberts, Barham, Chung, Sutton, Gehrmann et~al.}]{chowdhery2022palm}
Aakanksha Chowdhery, Sharan Narang, Jacob Devlin, Maarten Bosma, Gaurav Mishra,
  Adam Roberts, Paul Barham, Hyung~Won Chung, Charles Sutton, Sebastian
  Gehrmann, et~al. 2022.
\newblock Palm: Scaling language modeling with pathways.
\newblock \emph{arXiv preprint arXiv:2204.02311}.

\bibitem[{Desai and Durrett(2020)}]{calib_of_pretrained}
Shrey Desai and Greg Durrett. 2020.
\newblock Calibration of pre-trained transformers.
\newblock \emph{arXiv preprint arXiv:2003.07892}.

\bibitem[{Drozdov et~al.(2022)Drozdov, Sch{\"a}rli, Aky{\"u}rek, Scales, Song,
  Chen, Bousquet, and Zhou}]{drozdov2022compositional}
Andrew Drozdov, Nathanael Sch{\"a}rli, Ekin Aky{\"u}rek, Nathan Scales, Xinying
  Song, Xinyun Chen, Olivier Bousquet, and Denny Zhou. 2022.
\newblock Compositional semantic parsing with large language models.
\newblock \emph{arXiv preprint arXiv:2209.15003}.

\bibitem[{Fader et~al.(2011)Fader, Soderland, and
  Etzioni}]{fader2011identifying}
Anthony Fader, Stephen Soderland, and Oren Etzioni. 2011.
\newblock Identifying relations for open information extraction.
\newblock In \emph{Proceedings of the 2011 conference on empirical methods in
  natural language processing}, pages 1535--1545.

\bibitem[{Fellbaum(2020)}]{wordnet}
Christiane Fellbaum. 2020.
\newblock \emph{WordNet: An Electronic Lexical Database}.
\newblock MIT Press.

\bibitem[{Gao et~al.(2020)Gao, Biderman, Black, Golding, Hoppe, Foster, Phang,
  He, Thite, Nabeshima, Presser, and Leahy}]{pile}
Leo Gao, Stella Biderman, Sid Black, Laurence Golding, Travis Hoppe, Charles
  Foster, Jason Phang, Horace He, Anish Thite, Noa Nabeshima, Shawn Presser,
  and Connor Leahy. 2020.
\newblock The {P}ile: An 800gb dataset of diverse text for language modeling.
\newblock \emph{arXiv preprint arXiv:2101.00027}.

\bibitem[{Hao et~al.(2022)Hao, Tan, Tang, Zhang, Xing, and Hu}]{bertnet}
Shibo Hao, Bowen Tan, Kaiwen Tang, Hengzhe Zhang, Eric~P Xing, and Zhiting Hu.
  2022.
\newblock Bertnet: Harvesting knowledge graphs from pretrained language models.
\newblock \emph{arXiv preprint arXiv:2206.14268}.

\bibitem[{Jiang et~al.(2021)Jiang, Araki, Ding, and
  Neubig}]{jiang-etal-2021-know}
Zhengbao Jiang, Jun Araki, Haibo Ding, and Graham Neubig. 2021.
\newblock \href {https://doi.org/10.1162/tacl_a_00407} {How can we know when
  language models know? on the calibration of language models for question
  answering}.
\newblock \emph{Transactions of the Association for Computational Linguistics},
  9:962--977.

\bibitem[{Kadavath et~al.(2022)Kadavath, Conerly, Askell, Henighan, Drain,
  Perez, Schiefer, Dodds, DasSarma, Tran-Johnson
  et~al.}]{LM_know_what_they_know}
Saurav Kadavath, Tom Conerly, Amanda Askell, Tom Henighan, Dawn Drain, Ethan
  Perez, Nicholas Schiefer, Zac~Hatfield Dodds, Nova DasSarma, Eli
  Tran-Johnson, et~al. 2022.
\newblock Language models (mostly) know what they know.
\newblock \emph{arXiv preprint arXiv:2207.05221}.

\bibitem[{Khot et~al.(2022)Khot, Trivedi, Finlayson, Fu, Richardson, Clark, and
  Sabharwal}]{khot2022decomposed}
Tushar Khot, Harsh Trivedi, Matthew Finlayson, Yao Fu, Kyle Richardson, Peter
  Clark, and Ashish Sabharwal. 2022.
\newblock Decomposed prompting: A modular approach for solving complex tasks.
\newblock \emph{arXiv preprint arXiv:2210.02406}.

\bibitem[{Lin et~al.(2022)Lin, Hilton, and
  Evans}]{express_uncertainty_in_words}
Stephanie Lin, Jacob Hilton, and Owain Evans. 2022.
\newblock Teaching models to express their uncertainty in words.
\newblock \emph{arXiv preprint arXiv:2205.14334}.

\bibitem[{Petroni et~al.(2019)Petroni, Rocktäschel, Lewis, Bakhtin, Wu, and
  Miller}]{lama}
Fabio Petroni, Tim Rocktäschel, Patrick Lewis, Anton Bakhtin, Yuxiang Wu, and
  Sebastian Miller, Alexander H.~Riedel. 2019.
\newblock Language models as knowledge bases?
\newblock \emph{EMNLP}.

\bibitem[{Raffel et~al.(2020)Raffel, Shazeer, Roberts, Lee, Narang, Matena,
  Zhou, Li, and Liu}]{raffel2020t5}
Colin Raffel, Noam Shazeer, Adam Roberts, Katherine Lee, Sharan Narang, Michael
  Matena, Yanqi Zhou, Wei Li, and Peter~J. Liu. 2020.
\newblock \href {http://jmlr.org/papers/v21/20-074.html} {Exploring the limits
  of transfer learning with a unified text-to-text transformer}.
\newblock \emph{Journal of Machine Learning Research}, 21(140):1--67.

\bibitem[{Razniewski et~al.(2021)Razniewski, Yates, Kassner, and
  Weikum}]{lms_as_kbs}
Simon Razniewski, Andrew Yates, Nora Kassner, and Gerhard Weikum. 2021.
\newblock Language models as or for knowledge bases.
\newblock \emph{arXiv preprint arXiv:2110.04888}.

\bibitem[{Roberts et~al.(2020)Roberts, Raffel, and Shazeer}]{closed-book_qa}
Adam Roberts, Colin Raffel, and Noam Shazeer. 2020.
\newblock How much knowledge can you pack into the parameters of a language
  model?
\newblock \emph{EMNLP}.

\bibitem[{Speer et~al.(2017)Speer, Chin, and Havasi}]{conceptnet}
Robyn Speer, Joshua Chin, and Catherine Havasi. 2017.
\newblock Conceptnet 5.5: An open multilingual graph of general knowledge.
\newblock 31(1).

\bibitem[{Tay et~al.(2022)Tay, Tran, Dehghani, Ni, Bahri, Mehta, Qin, Hui,
  Zhao, Gupta et~al.}]{tay2022transformer}
Yi~Tay, Vinh~Q Tran, Mostafa Dehghani, Jianmo Ni, Dara Bahri, Harsh Mehta, Zhen
  Qin, Kai Hui, Zhe Zhao, Jai Gupta, et~al. 2022.
\newblock Transformer memory as a differentiable search index.
\newblock \emph{arXiv preprint arXiv:2202.06991}.

\bibitem[{Varshney et~al.(2022)Varshney, Mishra, and
  Baral}]{investigating_selective_predictions}
Neeraj Varshney, Swaroop Mishra, and Chitta Baral. 2022.
\newblock Investigating selective prediction approaches across several tasks in
  iid, ood, and adversarial settings.
\newblock \emph{arXiv preprint arXiv:2203.00211}.

\bibitem[{Vo and Bagheri(2017)}]{vo2017open}
Duc-Thuan Vo and Ebrahim Bagheri. 2017.
\newblock Open information extraction.
\newblock \emph{Encyclopedia with semantic computing and Robotic intelligence},
  1(01):1630003.

\bibitem[{Vrande{\v{c}}i{\'c} and Kr{\"o}tzsch(2014)}]{vrandevcic2014wikidata}
Denny Vrande{\v{c}}i{\'c} and Markus Kr{\"o}tzsch. 2014.
\newblock Wikidata: a free collaborative knowledgebase.
\newblock \emph{Communications of the ACM}, 57(10):78--85.

\bibitem[{Vrandečić and Krötzsch(2014)}]{WikiData}
Denny Vrandečić and Markus Krötzsch. 2014.
\newblock \href
  {http://cacm.acm.org/magazines/2014/10/178785-wikidata/fulltext} {Wikidata: A
  free collaborative knowledge base}.
\newblock \emph{Communications of the ACM}, 57:78--85.

\bibitem[{Wei et~al.(2022)Wei, Wang, Schuurmans, Bosma, Chi, Le, and
  Zhou}]{wei2022chain}
Jason Wei, Xuezhi Wang, Dale Schuurmans, Maarten Bosma, Ed~Chi, Quoc Le, and
  Denny Zhou. 2022.
\newblock Chain of thought prompting elicits reasoning in large language
  models.
\newblock \emph{arXiv preprint arXiv:2201.11903}.

\bibitem[{Yates et~al.(2007)Yates, Banko, Broadhead, Cafarella, Etzioni, and
  Soderland}]{yates2007textrunner}
Alexander Yates, Michele Banko, Matthew Broadhead, Michael~J Cafarella, Oren
  Etzioni, and Stephen Soderland. 2007.
\newblock Textrunner: open information extraction on the web.
\newblock In \emph{Proceedings of Human Language Technologies: The Annual
  Conference of the North American Chapter of the Association for Computational
  Linguistics (NAACL-HLT)}, pages 25--26.

\end{thebibliography}
\bibliographystyle{acl_natbib}

\appendix

\section{Technical Details}


\subsection{Relation-Paraphrasing}
\label{relation_para_extended_details}
We use 3 different instructions that have been manually constructed. If we denote a specific relation by $r$, then they are:
\begin{itemize}
    \item {\fontfamily{qcr}\selectfont"`$r$' may be described as"}
    \item {\fontfamily{qcr}\selectfont"`$r$' refers to"}
    \item {\fontfamily{qcr}\selectfont"please describe `$r$' in a few words:"}
\end{itemize}
That is, for every original relation which has been generated by the model, we perform additional three different model calls, one with each of those instruction prompts, resulting in three paraphrases. If needed, we eliminate overlapping paraphrases. 

\section{Full Prompts}

\subsection{Relation Generation}
\label{appen:relation_prompt}

{\fontfamily{qcr}\selectfont
\emph{Q}: Javier Culson 
\hfill \break
\emph{A}: participant of \# place of birth \# sex or gender \# country of citizenship \# occupation \# family name \# given name \# educated at \# sport \# sports discipline competed in 
\hfill \break
\hfill \break
\emph{Q}: René Magritte 
\hfill \break
\emph{A}: ethnic group \# place of birth \# place of death \# sex or gender \# spouse \# country of citizenship \# member of political party \# native language \# place of burial \# cause of death \# residence \# family name \# given name \# manner of death \# educated at \# field of work \# work location \# represented by 
\hfill \break
\hfill \break
\emph{Q}: Nadym 
\hfill \break
\emph{A}: country \# capital of  \# coordinate location \# population \# area \# elevation above sea level
\hfill \break
\hfill \break
\emph{Q}: Stryn 
\hfill \break
\emph{A}: significant event \# head of government \# country \# capital \# separated from 
\hfill \break
\hfill \break
\emph{Q}: 1585 
\hfill \break
\emph{A}: said to be the same as \# follows 
\hfill \break
\hfill \break
\emph{Q}: Bornheim 
\hfill \break
\emph{A}: head of government \# country \# member of \# coordinate location \# population \# area \# elevation above sea level
\hfill \break
\hfill \break
\emph{Q}: Aló Presidente 
\hfill \break
\emph{A}: genre \# country of origin \# cast member \# original network
}

\subsection{Pure Object Generation}
\label{appen:pure_obj_prompt}

{\fontfamily{qcr}\selectfont
\emph{Q}: Kristin von der Goltz \# mother 
\hfill \break
\emph{A}: Kirsti Hjort
\hfill \break
\hfill \break
\emph{Q}: Monte Cremasco \# country 
\hfill \break
\emph{A}: Italy
\hfill \break
\hfill \break
\emph{Q}: Johnny Depp \# children 
\hfill \break
\emph{A}: Jack Depp \# Lily-Rose Depp
\hfill \break
\hfill \break
\emph{Q}: Theodor Inama von Sternegg \# place of birth 
\hfill \break
\emph{A}: Augsburg
\hfill \break
\hfill \break
\emph{Q}: Wolfgang Sauseng \# employer 
\hfill \break
\emph{A}: University of Music and Performing Arts Vienna 
\hfill \break
\hfill \break
\emph{Q}: Hans Ertl \# sport 
\hfill \break
\emph{A}: mountaineering
\hfill \break
\hfill \break
\emph{Q}: Nicolas Cage \# sibling 
\hfill \break
\emph{A}: Christopher Coppola \# Marc Coppola
\hfill \break
\hfill \break
\emph{Q}: Manfred Müller \# occupation 
\hfill \break
\emph{A}: Catholic priest
}

\subsection{DK Object Generation}
\label{appen:dk_obj_prompt}

{\fontfamily{qcr}\selectfont
\emph{Q}: Heinrich Peters \# occupation 
\hfill \break
\emph{A}: Don't know
\hfill \break
\hfill \break
\emph{Q}: Monte Cremasco \# country 
\hfill \break
\emph{A}: Italy
\hfill \break
\hfill \break
\emph{Q}: Nicolas Cage \# sibling 
\hfill \break
\emph{A}: Christopher Coppola \# Marc Coppola
\hfill \break
\hfill \break
\emph{Q}: Hans Ertl \# sport 
\hfill \break
\emph{A}: mountaineering
\hfill \break
\hfill \break
\emph{Q}: Klaus Baumgartner \# work location 
\hfill \break
\emph{A}: Don't know 
\hfill \break
\hfill \break
\emph{Q}: Ruth Bader Ginsburg \# educated at 
\hfill \break
\emph{A}: Cornell University \# Harvard Law School \# Columbia Law School
\hfill \break
\hfill \break
\emph{Q}: Ferydoon Zandi \# place of birth 
\hfill \break
\emph{A}: Don't know
\hfill \break
\hfill \break
\emph{Q}: Wolfgang Sauseng \# employer 
\hfill \break
\emph{A}: University of Music and Performing Arts Vienna
\hfill \break
\hfill \break
\emph{Q}: Apayao \# head of government 
\hfill \break
\emph{A}: Don't know
\hfill \break
\hfill \break
\emph{Q}: Kristin von der Goltz \# mother 
\hfill \break
\emph{A}: Don't know
}

\section{Main Test Set}

\label{apx:main_test_set}

Table \ref{tab:main_test_set_seeds} provides our main test, which includes 100 different seeds - 4 from each of our predefined entity group categories.

\begin{table*}[t]
\setlength{\belowcaptionskip}{-10pt}
\footnotesize
\begin{center}
\begin{tabular}{l | l | l | l}
\textbf{Categories}  &\textbf{Sampled Seeds}  &\textbf{Categories}  &\textbf{Sampled Seeds}
\\ \toprule
\textbf{Politicians}   &  \makecell{Wang Zhi \\ Cathy Rogers \\ Kate Wilkinson \\ Carles Campuzano} & \textbf{Producers}   &  \makecell{Alyssa Milano \\ Lenny Kravitz \\ Carter Harman \\ Nancy Meyers}  \\ 
\hline

\textbf{Scientists}   &  \makecell{Pavel Krotov \\ Mirra Moiseevna Gukhman \\ Axel Delorme \\ Jesús Caballero Mellado}   & 
\textbf{Actors}  & \makecell{Jon Voight \\ Boris Savchenko \\ Tolga Tekin \\ Virginia Keiley} \\
\hline

\textbf{Basketball}   &  \makecell{Tom McMillen \\ Pat Kelly \\ Steve Moundou-Missi \\ Allen Phillips}   &
\textbf{Singers}        &  \makecell{Freddie Mercury \\ Angélique Kidjo \\ Camille Thurman \\ Giorgio Ronconi}   \\
\hline

\textbf{Sports}   &  \makecell{Peteca \\ motorcycle racing \\ Basque pelota \\ mountain bike trials}  &
\textbf{Bands}  &   \makecell{Steve Miller Band \\ Hypocrisy \\ Afro Kolektyw \\ Frailty} \\
\hline

\textbf{Artists } &  \makecell{Bořek Šípek \\ Loriot \\ George William Wakefield \\ George Trosley}  & 
\textbf{TV Shows}   &   \makecell{Secrets and Lies \\ Spirited Away \\ Super Friends \\ The Life and Legend of Wyatt Earp} \\
\hline

\textbf{Paintings}  &  \makecell{Portrait of a Man \\ Landscape \\ The foot washing \\ The King's rival} &
\textbf{Foods/Restaurants}  &  \makecell{Tahu petis \\ Kandil simidi \\ Jim Block \\ kubang boyo} \\
\hline

\textbf{Writers}   &  \makecell{Aleksandr Volkov \\ Osamu Tezuka \\ Elizaveta Sergeevna Danilova \\ Henry Saint Clair Wilkins}  &
\textbf{Animals}   &  \makecell{donkey \\ jaguar \\ mustang \\ whale} \\
\hline

\textbf{Books }  &  \makecell{The Green Berets \\ Alfred de Musset \\ Demain le capitalisme \\ The labyrinth}  &
\textbf{Plants}  &  \makecell{maple \\ rose \\ catmint \\ conflower} \\
\hline

\textbf{Landmarks}   &  \makecell{Trafalgar Square \\ Mount Everest \\ Yosemite National Park \\ Matterhorn}    & 
\textbf{Architects}     &  \makecell{Louis Kahn \\ Christopher Wren \\ Michael Graves \\ Domenico Fontana}\\
\hline

\textbf{Cities}   &  \makecell{Vatican City \\ Cherdyn \\ Toulon \\ MiljøXpressen}   &
\textbf{Drummers}    &  \makecell{Alan Montagu-Stuart-Wortley-Mackenzie \\ Mihály Deák \\ Joey Kramer \\ Stephanie Eulinberg} \\
\hline

\textbf{Countries}    &  \makecell{Niger \\ Sweden \\ England \\ Singapore}  &
\textbf{Biologists}   &  \makecell{Wangari Muta Maathai \\ James Rothman \\ Joanna Siódmiak \\ Barbara Bajd} \\
\hline

\textbf{Philosophy}   &  \makecell{Evgeny Torchinov \\ Nikolay Umov \\ Monica Giorgi \\ Larysa Tsitarenka} & 
\textbf{AI Researchers}     &  \makecell{William T. Freeman \\ Stephen Falken \\ Joseph Weizenbaum \\ Robby Garner} \\
\hline

\textbf{Movies}    &  \makecell{Spider-Man: Far from Home \\ Sonic the Hedgehog \\ Unearthed \\ Another Man's Poison}  \\

\bottomrule
\end{tabular}
\caption{List of all main test set seeds }
\label{tab:main_test_set_seeds}
\end{center}
\end{table*}

\section{Automatic Precision Evaluation}
\label{apx:automatic_eval}
As noted in the main text, the automatic precision evaluation method (i.e., the one based on Google search) may sometimes fail. Some of the failure cases are: (a) \emph{Inexact string matching}. For example \textsc{(Boston Celtics, league, National Basketball Association (NBA))} is not verified, but dropping \textsc{(NBA)} from the object would result in a successful verification. b) \emph{Paraphrases}: For example \textsc{(Marble Arch, country, United Kingdom)} is not verified but changing the object to \textsc{England} does succeed.

\section{Additional Generated Graphs}
\label{apx:example_graphs}

\begin{figure*}
    \centering
    \includegraphics[scale=0.45]{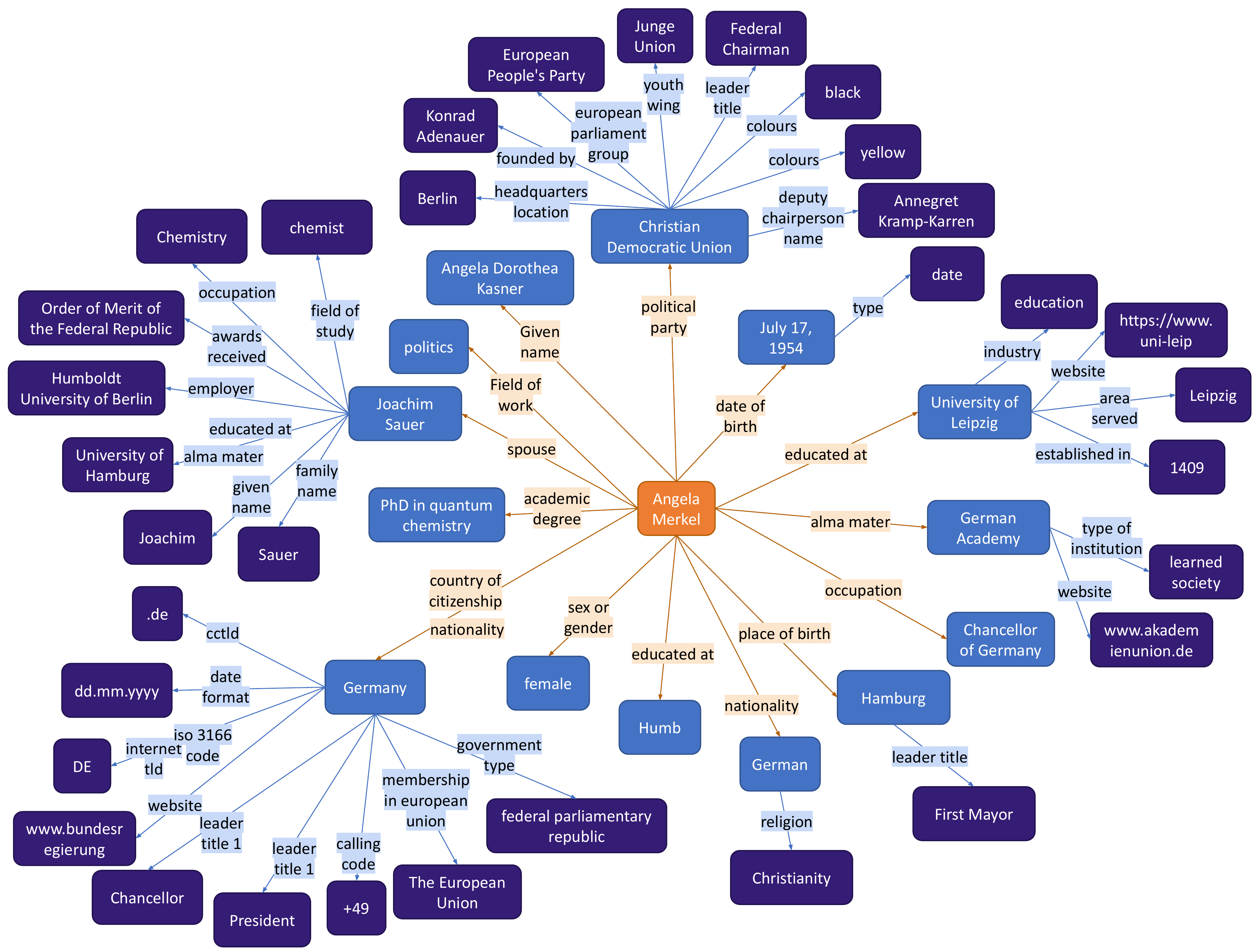}
    \caption{An example of a generated depth-2 knowledge graph around the seed entity \textsc{Angela Merkel}, using \OP (see \secref{sec:crawling}). For readability, back edges from 2-depth nodes to 1-depth nodes are omitted.}
    \label{fig:angela_markel}
\end{figure*}

\begin{figure*}
    \centering
    \includegraphics[scale=0.45]{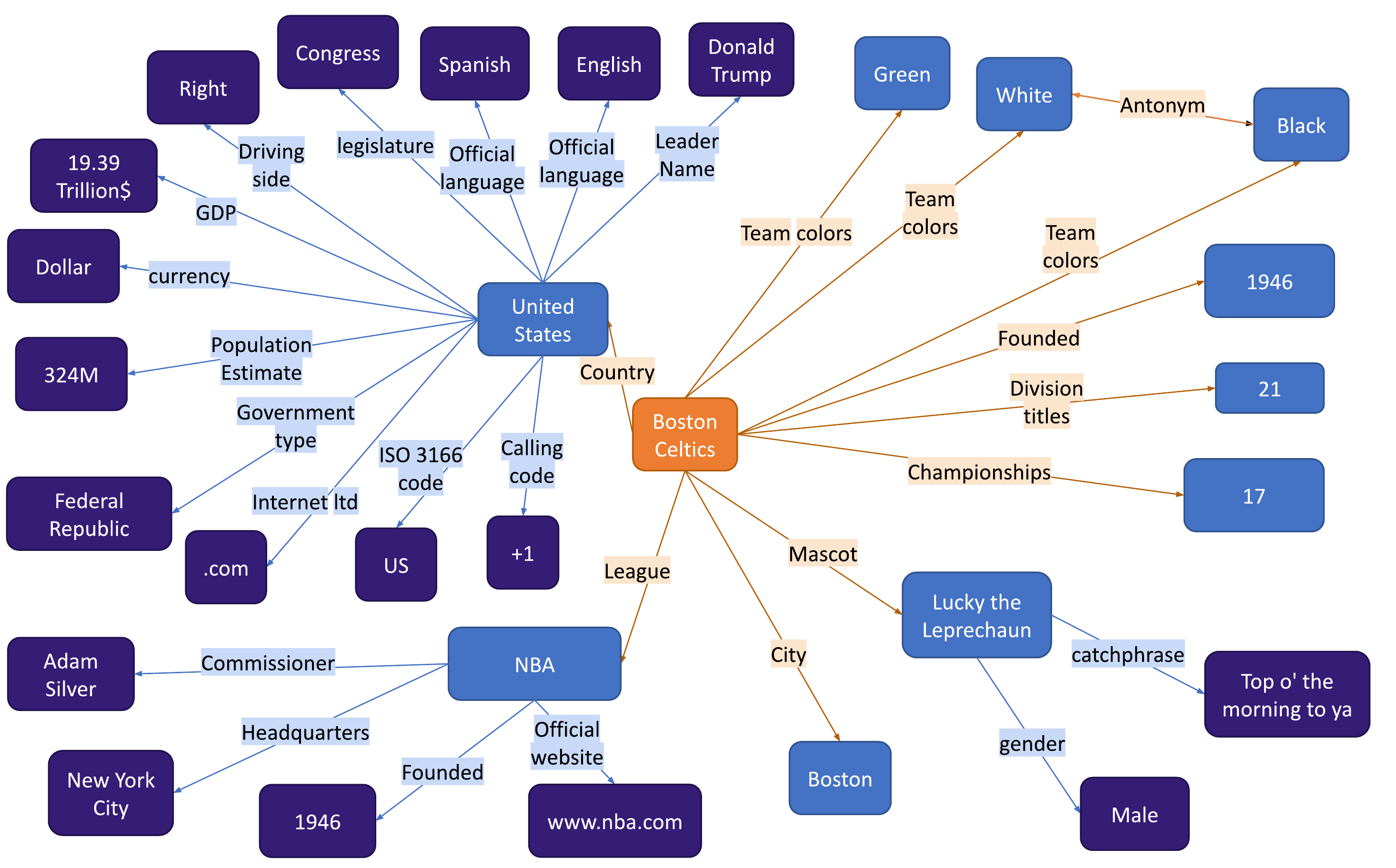}
    \caption{An example of a generated depth-2 knowledge graph around the seed entity \textsc{Boston Celtics}, using {\OP} (see \secref{sec:crawling}).}
    \label{fig:boston_celtics}
\end{figure*}

Figs.~\ref{fig:angela_markel}, ~\ref{fig:boston_celtics} show additional example graphs (to the one shown in Fig.~\ref{fig:knolwedge_graph}), generated around the seed entities \textsc{Angela Merkel} and \textsc{Boston Celtics} respectively.

\end{document}